\documentclass[runningheads]{llncs}
\usepackage{graphicx}
\usepackage{url}
\usepackage{amsmath,amssymb} 
\usepackage{color}
\usepackage{xspace}

\newcommand*{\ie}{\emph{i.e.}\@\xspace}

\newcommand*{\etc}{\emph{etc.}\@\xspace}

\begin{document}
\title{RPNet: an End-to-End Network for Relative Camera Pose Estimation} 

\titlerunning{RPNet: an End-to-End Network for Relative Camera Pose Estimation}
%
\author{Sovann En \and Alexis Lechervy \and Fr\'ed\'eric Jurie }
%
\authorrunning{S. En, A. Lechervy, F. Jurie}
%

\institute{Normandie Univ, UNICAEN, ENSICAEN, CNRS --- UMR GREYC
\email{firstname.lastname@unicaen.fr}
}
\maketitle              
\begin{abstract}
This paper addresses the task of relative camera pose estimation from raw image pixels, by means of deep neural networks. The proposed RPNet network takes pairs of images as input and directly infers the relative poses, without the need of camera intrinsic/extrinsic. While state-of-the-art systems based on SIFT + RANSAC, are able to recover the translation vector only up to scale, RPNet is trained to produce the full translation vector, in an end-to-end way. Experimental results on the Cambridge Landmark data set show very promising results regarding the recovery of the full translation vector. They also show that RPNet produces more accurate and more stable results than traditional approaches, especially for hard images (repetitive textures, textureless images, \etc). To the best of our knowledge, RPNet is the first attempt to recover full translation vectors in relative pose estimation.

\keywords{relative pose estimation \and pose estimation \and posenet}
\end{abstract}
\section{Introduction}
\label{sec:introduction}
In this paper, we are interested in {\em relative camera pose estimation} --- a task consisting in accurately estimating the location and orientation of the camera with respect to another camera's reference system. Relative pose estimation is an essential task for many computer vision problems, such as Structure from Motion (SfM), Simultaneous Localisation And Mapping (SLAM), \etc Traditionally, this task can be accomplished by i) extracting sparse keypoints (ex. SIFT, SURF), ii) establishing 2D correspondences between keypoints and iii) estimating the essential matrix using 5-points or 8-point algorithms \cite{nister2004efficient}. RANSAC is very often used to reject outliers in a robust manner. 

This technique, although it has been considered as the de facto standard for many years, presents two main drawbacks. First, the quality of the estimation depends heavily on the correspondence assignment. This is to say, too few correspondences (textureless objects) or too many noisy correspondences (repetitive texture or too much viewpoint change) can lead to surprisingly bad results. 
Second, the traditional method is able to estimate the translation vector only up to scale (directional vector).

In this paper, our objective is three folds: i) we propose a system producing more stable results ii)  recovering the full translation vector iii) and we provide insights regarding relative pose estimation (\ie. from absolute pose, from a pose regressor \etc). 

As pointed out in \cite{walch2016image}, CNN based methods are able to produce pretty good results in some cases where SIFT-based methods fail (\ie texture less images). This is the reason why we opted for a global method based on CNN. Inspired by the success of PoseNet \cite{kendall2015posenet}, we propose a modified Siamese PoseNet for relative camera pose estimation, dubbed as RPNet, with different ways to infer the relative pose. To the best of our knowledge, \cite{melekhov2017relative} is the only end-to-end system aiming at solving relative camera pose using deep learning approach. However, their system estimate the translation vector up to scale, while ours produces full translation vectors.

The rest of the paper is organized as follows: Section \ref{sec:state_of_the_art} presents the related work. Section \ref{sec:system} introduces the network architecture and the training methodology. Section \ref{sec:experimentation} discusses the datasets and presents the experimental validation of the approach. Finally, Section 5 concludes the paper.

\section{State of the Art}
\label{sec:state_of_the_art}

\textbf{Local keypoint-based approaches.} They address relative camera pose estimation using the epipolar geometry between 2D-2D correspondences of keypoints. 
Early attempts aimed at better engineering interest point detectors to focus on interesting image properties such as corners \cite{harris1988combined}, blobs in scale-space \cite{lowe2004distinctive}, regions \cite{matas2004robust}, or speed \cite{bay2006surf,tola2010daisy,rublee2011orb} \etc More recently, there is a growing interest to train interest point detectors together with the matching function \cite{han2015matchnet,zagoruyko2015learning,tian2017l2,detone2017superpoint,trujillo2006using}. LIFT \cite{yi2016lift} adopted the traditional pipeline combining a detector, an orientation estimator, and a descriptor, tied together with differentiable operations and learned end-to-end. \cite{altwaijry2016learning} proposed a multitask network with different sub-branches to operate on varying input sizes. \cite{detone2017superpoint} proposed a bootstrapping strategy by first learning on simple synthetic data 
and increasing the training set with real images in a second time.

\noindent \textbf{End-to-End pose estimation.} The first end-to-end neural network for camera pose estimation from single RGB images is PoseNet \cite{kendall2015posenet}. It is based on GoogLeNet with two output branches to regress translations and rotations. PoseNet follow-up includes: Baysian PoseNet \cite{kendall2016modelling}, Posenet-LSTM \cite{walch2016image} where LSTM is used to model the context of the images, Geometric-PoseNet where the loss is calculated using the re-projection error of the coordinates using the predicted pose and the ground truth \cite{kendall2017geometric}. Since all the 3D models used for comparisons are created using SIFT-based techniques, traditional approach seems more accurate. \cite{walch2016image} showed that the classical approaches completely fail with less textured datasets such as the proposed TMU-LSI dataset. \cite{sppnet2018} is an end-to-end system for pose regression taking sparse keypoint as inputs. Regarding relative pose estimation, \cite{melekhov2017relative} is the only system we are aware of. Their network is based on ResNet35 with FCs layers acting as pose regressor. Similar to the previous networks, the authors formulate the loss function as minimising the L2-distances between the ground truth and the estimated pose. Unfortunately, several aspects of their results (including their label generation, experimental methodology and the baseline system) make comparisons difficult. Along side with pose regression problems, another promising works from \cite{rocco2017convolutional} showed that an end-to-end neural network can effectively be trained to regress to infer the homography between two images. Finally, two recent papers \cite{yi2018learning,brachmann2017dsac} made useful contributions to the training of end-to-end systems for pose estimation. \cite{yi2018learning} proposed a regressor network to produce essential matrix which can be then used to find the relative pose. However, their system is able to find the translation up to scale which is completely different from our objective. In \cite{brachmann2017dsac}, a differentiable RANSAC is proposed for outlier rejection and can be a plug-and-play component into an end-to-end system. 

\begin{figure}[tb]
\centering
\includegraphics[width=0.9\textwidth]{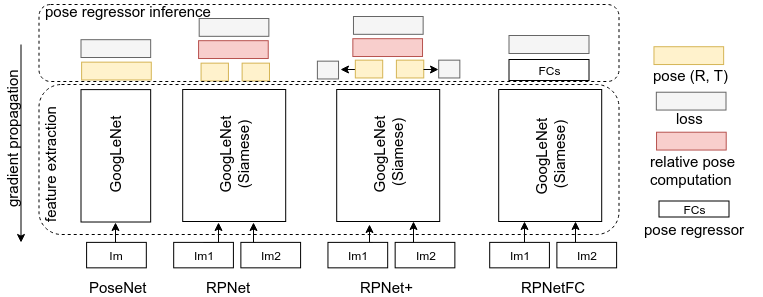}
\vspace{-1em}
\caption{Illustration of the proposed system}
\label{fig:system_architecture}
\vspace{-1em}
\end{figure} 

\section{Relative pose inference with RPNet}
\label{sec:system}

\noindent \textbf{Architecture.} The  architecture of the proposed RPNet, illustrated Fig. \ref{fig:system_architecture}, is made of two building blocks: i) a Siamese Network with two branches regressing one pose per image, ii) a pose inference module for computing the relative pose between the cameras. We provide three variants of the pose inference module: (1) a parameter-free module, (2) a parameter-free module with additional losses (same as PoseNet loss~\cite{kendall2015posenet}) aiming at regressing the two camera poses as well as the relative pose, and (3) a relative pose regressor based on FC layers. The whole network is trained end-to-end for relative pose estimation. Inspired by PoseNet \cite{kendall2015posenet}, the feature extraction network is based on the GoogLeNet architecture with 22 CNN layers and 6 inception modules. We only normalize the quaternion during test time. It outputs one pose per image.

For RPNet and RPNet$^{+}$, the module for computing the relative pose between the cameras is straightforward and relies on simple geometry.  Following the convention of OpenCV, the relative pose is calculated in the reference system of the 2nd camera. Let $(R_1, t_1, R_2, t_2)$ be the rotation matrices and translation vectors used to project a point $X$ from world coordinate system to a fixed camera system (camera 1 \& 2). Let $(q_1, q_2)$ be the corresponding quaternions of $(R_1, R_2)$. The relative pose is calculated as followed: \begin{equation}
R_{1,2} = q_2 \times q_1^* \quad\text{and}\quad T_{1,2} = R_2 (-R_1^Tt_1) + t_2
\label{equa:relative_translation}
\end{equation}
where $q_1^*$ is the conjugate of $q_1$, and $\times$ denotes the multiplication in the quaternion domain. Both equations are differentiable. For RPNetFC, the pose inference module is a simple stacked fully connected layers with $relu$ activation. To limit over-fitting, we modified the output of the Siamese network by reducing its output dimension from 2048 to 256. This results in almost 50\% reduction of the number of parameters compared to PoseNet, RPNet and RPNet$^{+}$ network. The pose regressor network contains two FC layers (both with 128 dimensions). 

\noindent \textbf{Losses.} The loss function uses the Euclidean distance to compare predicted relative rotation $\hat{q_{1,2}}$ and translation $\hat{T}_{1,2}$ with ground truth $\hat{q}_{1,2}$ and $q_{1,2}$ : $loss = \sum_i(||\hat{T}^i_{1,2} - T_{1,2}^i||_2 + \beta * ||\hat{q}_{1,2}^i - q_{1,2}^i||_2$). Quaternions are unit quaternions. The original PoseNet has a $\beta$ term in front of quaternions to balance the loss values between the translation and rotation. To find the most suitable value of $\beta$, we cross-validated on our validation set. Please refer to our codes for different hyper-parameter values on different subsets. 

\section{Experimentations}
\label{sec:experimentation}

\subsection{Experimental Setup}

\noindent \textbf{Dataset.} Experimental validation is done on the Cambridge Landmark dataset\footnote{\url{http://mi.eng.cam.ac.uk/projects/relocalisation}}. Each image is associated with a ground-truth pose. We provide results on 4 of the 5 subsets (scenes). As discussed by several people, the 'street' scene raises several issues\footnote{\url{https://github.com/alexgkendall/caffe-posenet/issues/2}}.

\begin{table}[tb]
      \centering
	    \caption{Number of training and testing pairs for Cambridge Landmark dataset. SE stands for spatial extent, measured in meter.}
        \begin{tabular}{l|l|l|l||l|l|l|l}
\hline
Scene        & Train & Test & SE     & Scence          & Train & Test & SE    \\ \hline
Kings College & 9.1k  & 2.4k & 140x40 & Shop Facade     & 1.6k  & 0.6k & 35x25 \\
Old Hospital & 6.5k  & 1.2k & 50x40  & St Marys Church & 11k  & 4.1k & 80x60 \\ \hline
		\end{tabular}
   \vspace{-1em}
\end{table}

\noindent \textbf{Pair generation.}   For each sequence of each scene, we randomly pair each image with eight different images of the same sequence. For a fair comparison with SURF, the pair generation is done by making sure that they overlap enough. We followed the train-test splits defined with the data set. Images are scaled so that the smallest dimension is 256 pixels, keeping its original aspect ratio. During training, we use 224*224 random crops and feed them into the network. During test time, we center crop the image. 

\noindent \textbf{Baseline.} The baseline is a traditional keypoint-based method (SURF). The focal length and the principle point are provided by the dataset. Other parameters are cross-validated on the validation set. For a fair comparison, we provide two scenarios for baselines: (1) the image are scaled to be 256*455 pixels, followed by a center-crop (224*224 pixels) to produce the same image pairs as tested with our networks and (2) the original images without down-sampling. We named these two scenarios as 'SURFSmall' and 'SURFFull'. All the camera parameters are adapted to the scaling and cropping we applied.

\noindent \textbf{Evaluation metric.} We measured 3 different errors: i) translation errors, in meters ii) rotation errors, in degrees and iii) translation errors in degrees. We report the median for all the measurements.

\subsection{Experimental results} 
\begin{figure}[tb]
\centering
\includegraphics[width=0.95\textwidth]{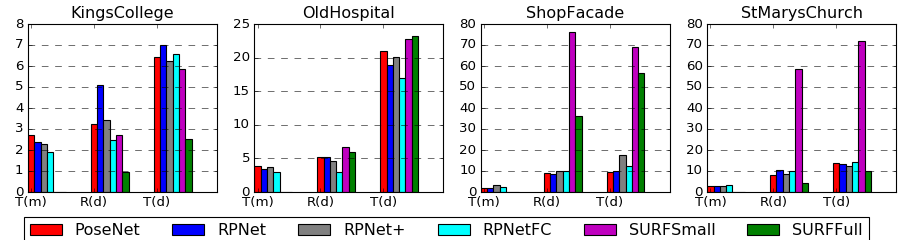}
\vspace{-1em}
\caption{Translation and Rotation errors (median) of the different approaches}
\label{fig:result_comparison}
\vspace{-1em}
\end{figure}

\noindent \textbf{Relative pose inference module.} Fig. \ref{fig:result_comparison} compares the performance of the different systems and test scenarios. Based on these experimental results, RPNetFC and RPNet$^{+}$ are the most efficient ways to recover the relative pose. On easy dataset (\ie. KingsCollege and OldHospital), where there is no ambiguity textures, using pose regressor (RPNetFC) produces slightly better results than inferring the relative pose from the two images (PoseNet/RPNet/RPNet$^+$). On the contrary, on hard datasets (\ie. ShopFacade and StMarysChurch), RPNet-family outperforms RPNetFC. This behavior is also true for relative rotation and relative translation measured in degree. Globally, RPNetFC produces the best results followed by RPNet$^+$, PoseNet and finally, RPNet. The differences of their results are between 0 and 8 degrees. Regarding technical aspect, RPNetFC is a lot easier to train than RPNet$^+$/RPNet since it does not involve multiple hyper-parameters to balance the different losses. It also converges faster.

\begin{figure}[tb]
\centering
\includegraphics[width=0.95\textwidth]{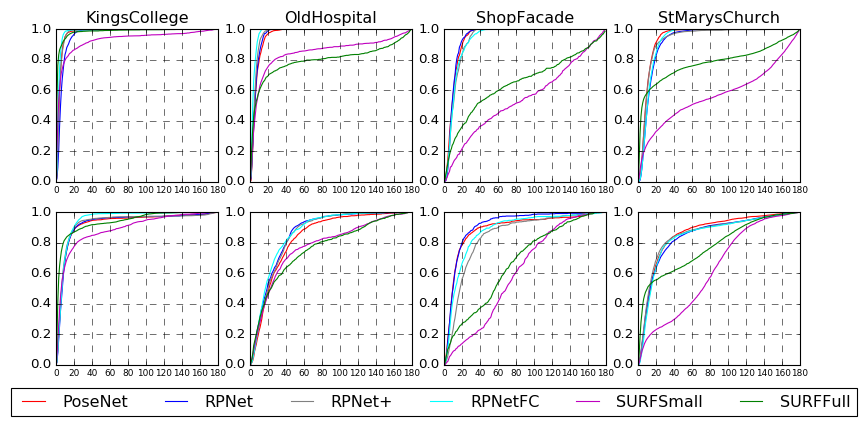}
\vspace{-1em}
\caption{Accumulative hist. of errors in rotation (1st row, d), translation (2nd row, m). }
\label{fig:accum_hist}
\vspace{-1em}
\end{figure}

\noindent \textbf{Comparison with traditional approaches.} We will start by discussing the SURFSmall scenario first. In general, the error on both translation and rotation can be reduced between 5 to 70\% using RPNet family, except on KingsCollege where the traditional approach slightly outperforms RPNet-based methods. We observed that the performance of the traditional approaches varies largely from one subset to another, while RPNet$^+$/RPNetFC are more stable. In addition, the traditional approach requires camera information for each image in order to correctly estimate the pose. In contrast, RPNet-based does not require any specific information at all. Using the original image size (SURFFull) significantly boost the performance of the traditional approach. However, RPNetFC still enjoy a significant gain in performance on OldHospital and ShopFacade, while performing slightly worse than SURFFull on KingsCollege and StMarysChurch. The difference in performance between SURFFull and RPNetFC is even more significant when the images contain large view point changes (see Fig. \ref{fig:accum_hist}). 

\begin{figure}[tb]
\centering
\includegraphics[width=0.95\textwidth]{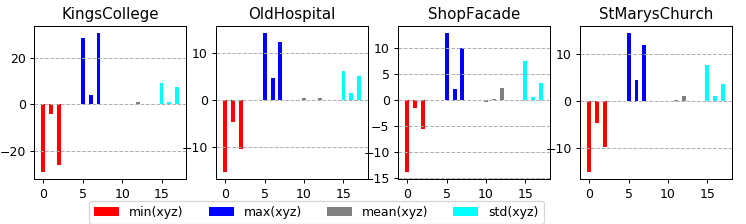}
\vspace{-1em}
\caption{Min/Max/Mean/STD relative translations (ground truth), w.r.t. XYZ axis (m).}
\label{fig:spatial_extent}
\vspace{-0.8em}
\end{figure}

\noindent \textbf{Full translation vector.} One of our objectives is to provide a system able to estimate the full translation vector. On average, we observed that the median error ranges between 2 to 4 meters, using RPNetFC. Fig. \ref{fig:spatial_extent} gives an idea of ground truth translations w.r.t. reference axes (xyz). For instance, on KingsCollege, the values of X-axis can range from -29m to 30m with an STD of 7 meters. Interestingly, our network has a translation error of only 2.88 meters. 

\section{Conclusions}

This paper proposed a novel architecture for estimating  full relative poses using an end-to-end trained neural network. The network is based on a Siamese architecture, which was experimented with different ways to infer the relative poses. In addition, to produce competitive or better results over the traditional SURF-based approaches, our system is able to produce an accurate full translation vector. We hope this paper will provide more insight and motivate other researchers to focus on global end-to-end system for relative pose regression problems.\\

\noindent \textbf{Acknowledgements}. This work was partly funded by the French–UK MCM ITP program and by the ANR-16-CE23-0006 program.
\clearpage

\bibliographystyle{splncs04}
\bibliography{egbib}

\end{document}